\documentclass[letterpaper]{article} 
\usepackage{aaai2026}  
\usepackage{times}  
\usepackage{helvet}  
\usepackage{courier}  
\usepackage[hyphens]{url}  
\usepackage{graphicx} 
\usepackage{natbib}  
\usepackage{caption} 
\usepackage{algorithm}
\usepackage{algorithmic}
\usepackage{multirow}
\usepackage{inputenc}
\usepackage{newfloat}
\usepackage{listings}
\DeclareCaptionStyle{ruled}{labelfont=normalfont,labelsep=colon,strut=off} 
\floatstyle{ruled}
\newfloat{listing}{tb}{lst}{}
\floatname{listing}{Listing}
\title{Harnessing the Unseen: The Hidden Influence of \\
Intrinsic Knowledge in Long-Context Language Models}
\author {
    Yu Fu\textsuperscript{\rm 1},
    Haz Sameen Shahgir\textsuperscript{\rm 1},
    Hui Liu\textsuperscript{\rm 2},
    Xianfeng Tang\textsuperscript{\rm 2},
    Qi He\textsuperscript{\rm 3},
    Yue Dong\textsuperscript{\rm 1}\thanks{Corresponding Author}
}
\affiliations {
    \textsuperscript{\rm 1}University of California, Riverside,\\
    \textsuperscript{\rm 2}Amazon,\\
    \textsuperscript{\rm 3} Microsoft \\
    yfu093@ucr.edu, yue.dong@ucr.edu
}

\begin{document}

\maketitle

\begin{abstract}
Recent advances in long-context language models (LCLMs), designed to handle extremely long contexts, primarily focus on utilizing external contextual information, often leaving the influence of language models' parametric knowledge underexplored. In this work, we firstly investigate how this parametric knowledge affects content generation and demonstrate that its impact becomes increasingly pronounced as context length extends. Furthermore, we show that the model’s ability to utilize parametric knowledge, which we call parametric recall ability, does not improve simultaneously with its ability to leverage contextual knowledge through extrinsic retrieval ability. Moreover, better extrinsic retrieval ability can interfere with the model’s parametric recall ability, limiting its full potential.  To bridge this gap, we design a simple yet effective Hybrid Needle-in-a-Haystack test that evaluates models based on their capabilities across both abilities, rather than solely emphasizing extrinsic retrieval ability. Our experimental results reveal that Qwen-2.5 models significantly outperform Llama-3.1 models, demonstrating superior potential to combine various abilities. Moreover, even the more powerful \texttt{Llama-3.1-70B-Instruct} model fails to exhibit better performance, highlighting the importance of evaluating models from a dual-ability perspective.
\end{abstract}

\begin{links}
    \link{Code}{https://github.com/FYYFU/Hybrid-NIAH}
\end{links}

\section{Introduction}

Recent advancements in both open-source models (e.g., LLaMA~\citep{llama3}, Qwen~\citep{qwen2, qwen2.5}) and closed-source models (e.g., GPT-4~\citep{gpt4, gpt4o}, Claude~\citep{claude3}, Gemini~\citep{gemini2}) have incorporated long-context capabilities and significantly extended their context windows. For instance, GPT-4o and Claude Sonnet 4 have expanded their context windows to 128K and 200K tokens respectively. The emergence of those Long Context Language Models (LCLMs) has significantly advanced the ability of language models to process and generate coherent content over extended contexts. They have become increasingly useful for various tasks, including document summarization~\citep{yen2025helmet_evaluate}, multi-turn conversations~\citep{long_convo_mem} and question answering~\citep{karpinska2024_thousandpairs}, where retaining and utilizing long-term dependencies is essential. 

With the advance of LCLMs, numerous benchmarks have been proposed to evaluate their effectiveness in handling long contexts. For example, the Needle-in-a-Haystack test~\citep{niah} inserts fabricated critical information (i.e. ``needle") into long irrelevant documents (i.e. ``haystack") to examine LCLMs' external retrieval ability and prevent the influence of their own parametric knowledge. Other benchmarks such as RULER~\citep{hsieh2024_RULER}, InfiniteBench~\citep{infinity-bench} and HELMET~\citep{yen2025helmet_evaluate} extend to more realistic tasks such as QA, summarization etc., and ultra long contexts (100K+ tokens) to systematically assess LCLMs' capacity to process and generate meaningful outputs over extraordinarily external long contexts. These benchmarks evaluate how effectively these models leverage external long contexts, thereby emphasizing their extrinsic retrieval ability while overlooking the role of parametric knowledge~\citep{how-much-knowledge, knowledge-base-for-kb, what-lms-know}. 

However, the parametric knowledge has already been shown to be pivotal to the performance of language models, especially when parametric and extrinsic knowledge contradict each other~\citep{yu_su_knowledge_conflict,wang2024astuteragovercomingimperfect,knowledge_conflict_survey}. \textbf{In this work, we argue that parametric knowledge plays an important role during long-context generation, with its impact becoming more pronounced as context length extends.}  To validate this hypothesis, we construct a new dataset, \texttt{IWhoQA}, based on WhoQA \citep{WhoQA}, specifically designed to systematically probe LCLMs in scenarios where parametric knowledge either aligns with or contradicts the external context. Experimental results show that LCLMs consistently perform better when the external context aligns with their parametric knowledge. The performance gap increases with longer contexts, reaching up to 10 points on \texttt{Llama-3.1-8B-Instruct}, suggesting a growing reliance on parametric knowledge during long-context generation.

Given the importance of parametric knowledge for LCLMs and the lack of its evaluation in existing benchmarks, it remains unclear whether methods that perform well in these benchmarks improve only extrinsic retrieval ability, or also enhance parametric recall ability. To address this gap, we systematically investigate the relationship between extrinsic retrieval and parametric recall. Specifically, we compare RoPE~\citep{Su2021RoFormerET} and STRING~\citep{string_pe}, an improved variant of RoPE designed for long-context tasks, across various curated datasets featuring diverse knowledge alignments between external context and parametric knowledge. Our results show that while STRING improves extrinsic retrieval ability, it suppresses parametric recall even when parametric knowledge benefits generation. This reveals a trade-off between leveraging external context and utilizing parametric knowledge, an aspect overlooked by current benchmarks.

Building on these observations, we propose a simple yet effective Hybrid Needle-in-a-Haystack test to comprehensively evaluate how well models integrate parametric recall ability with extrinsic retrieval ability during long-context generation. Specifically, we design queries such as \textit{``What’s the favorite thing of the person who wrote \{Book\_Name\}?''}—which require the model to first recall the author’s name from its parametric knowledge, and then retrieve the inserted needle in the context to answer the question. Experimental results show that the Qwen2.5 family demonstrates a near-linear improvement in its ability to integrate both types of knowledge as model size increases. In contrast, the Llama3.1 family exhibits minimal gains despite a substantial increase in parameters, suggesting that it struggles to effectively leverage parametric knowledge in long-context settings. Our contributions can be summarized as follows:
\begin{itemize}

    \item We present the first study demonstrating that the parametric knowledge of LCLMs plays an important role in long-context generation, with its influence becoming increasingly pronounced as the context length extends.
    
    \item We show that naively improving LCLMs' extrinsic retrieval ability will unintentionally suppress their parametric recall ability, revealing a trade-off between utilizing external contexts and  leveraging parametric knowledge.

    \item We introduce a simple yet effective Hybrid Needle-in-a-Haystack test to examine LCLMs' ability to integrate the parametric recall ability and extrinsic retrieval ability during long-context generation. We show that this ability doesn't always scale with the model size, highlighting limitations in existing LCLMs.
\end{itemize}

\section{Related Work}

\paragraph{Parametric Knowledge}
During pre-training, LLMs memorize a vast amount of knowledge into their parameters, known as parametric (intrinsic)  knowledge \citep{how-much-knowledge, knowledge-base-for-kb, what-lms-know}. This parametric knowledge can often conflict with new information presented via prompting through a phenomenon known as knowledge conflict \citep{yu_su_knowledge_conflict,entity_based_kc,misinformation_attack, persuade-llm}. While earlier works have found that LLMs rely on parametric knowledge when prompted with simple counter-factual statements \citep{entity_based_kc, rich_source}, more recent research has demonstrated that LLMs rely more on the information presented as the context \citep{yu_su_knowledge_conflict, misinformation_attack, convincing-evidence}. \citet{misinformation_attack} uncovers LLMs' sensitivity to small amounts of even misinformation in context and \citet{persuade-llm} applies prompting to persuade LLMs to disregard their parametric knowledge. However, \citet{yu_su_knowledge_conflict} further finds that when presented with conflicted evidence, LLMs resolve in favor of parametric knowledge. \citet{context_faithful_prompting} proposes framing the conflicting information as user opinion to prevent the LLM from taking it as fact. We refer the reader to \citet{knowledge_conflict_survey} for further details. However, to the best of our knowledge, the relation between parametric knowledge and long-context tasks has not been studied.

\paragraph{Long Context Models}

Training transformers on long context from scratch is prohibitively expensive. Much research has focused on extending the context window of existing models through two categories: 1) Finetuning on a small amount of long context data \citep{peng2024yarn, cerebras2025longcontext, position_interpolation_training, ntkbyparts} and 2) Training-free methods based on reusing position information. \citep{dual_chunk_attention, NTKAware2023, DynamicNTK, rerope2023, self-extend}. For example, Self-Extend \citep{self-extend}, Dual Chunk Attention \citep{dual_chunk_attention}, and STRING \citep{string_pe} all propose replacing large relative positions by overwriting them with proportionately smaller ones seen during training. By mapping large relative positions to smaller ones that are well represented during pre-training, the model can more effectively integrate provided information into long-context generation.
\paragraph{Long Context Evaluation}
Multiple synthetic tasks have been proposed to evaluate LCLMs, including Needle-in-a-Haystack (NIAH) test~\citep{niah},  long-context QA, summarization, reasoning, extrinsic learning (ICL) and retrieval-augmented generation (RAG), among others \citep{hsieh2024_RULER, kim2025_oneruler, gao2025_uniah, liu2024_longgenbench, yen2025helmet_evaluate}. Variants of NIAH include multiple-needle, long-needle, and needle-in-needle configurations \citep{gao2025_uniah} and have also been extended to test long-context reasoning~\citep{kuratov2024babilong}, mathematical~\citep{wang2024mathhay_niah} and multilingual capabilities \citep{multilingual_niah}. However, several works have found that the standard NIAH task may not truly reflect an LCLM's downstream capabilities and have proposed comprehensive benchmarks such as RULER \citep{hsieh2024_RULER} and its multilingual extension ONERULER \citep{kim2025_oneruler}, LongBench \citep{longbench} , LongGenBench~\citep{liu2024_longgenbench}, L-Eval \citep{an2023leval}, InfiniteBench~\citep{infinity-bench} and HELMET \citep{yen2025helmet_evaluate}. These benchmarks contain several tasks such as citation generation, reranking, QA, summarization, ICL, and RAG. \citet{karpinska2024_thousandpairs} introduced NoCha, a dataset of human-written true and false claims about recently published fictional books. While previous work has primarily studied long context in isolation, some going so far as to use entirely fictional contexts \citep{gao2025_uniah, karpinska2024_thousandpairs}, to the best of our knowledge the problem of how an LCLM's parametric knowledge affects its long context abilities has not been explored.

\begin{figure*}[t]
    \centering
    \includegraphics[width=0.81\textwidth]{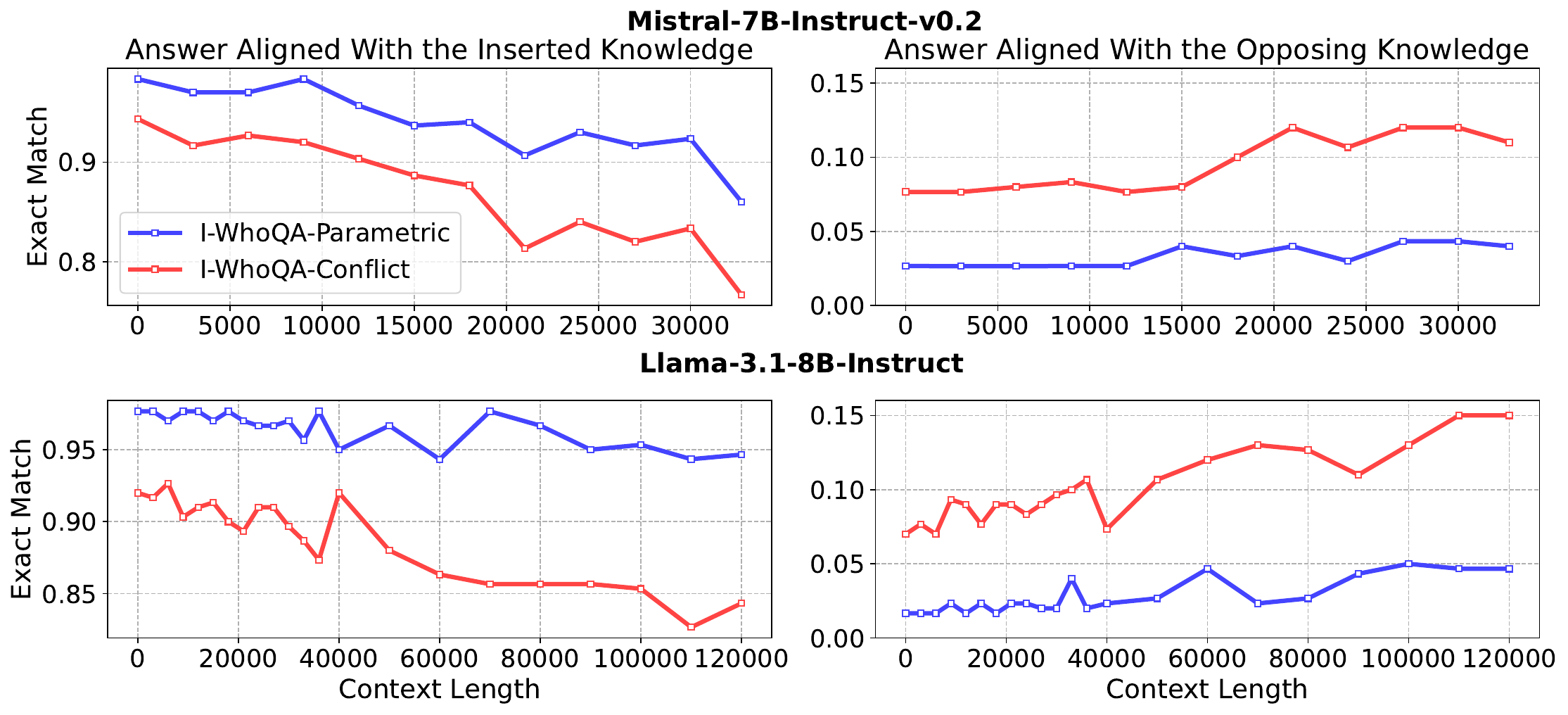}

    \caption{ We only count answers that align with either the injected knowledge (whether parametric or conflictual) or the opposing knowledge source. Left) LCLMs struggle to retrieve the answer from the context on the I-WhoQA-Conflict subset, i.e., when the context information conflicts with their parametric knowledge. Right) The upward trend of I-WhoQA-conflict (red) shows that when the parametric knowledge conflicts with the context, the likelihood of a LCLM relying on parametric knowledge steadily increases with larger contexts. 
    }
    \label{fig:section1}
\end{figure*}

\section{The Role of Parametric Knowledge in Long-Context Generation}
\label{point1}

Existing benchmarks for evaluating LCLMs primarily focus on assessing their extrinsic retrieval ability—how well they extract information or perform reasoning based solely on the external context. In contrast, our work aims to investigate the role of parametric knowledge embedded within LCLMs. We begin by posing a central question: \textbf{What role does the parametric knowledge of LCLMs play in long-context generation?} Specifically, we explore how parametric knowledge interacts with external context and how this interplay affects the quality of the generated output.

\subsection{Dataset Creation}
While existing datasets in the knowledge conflict~\citep{yu_su_knowledge_conflict, knowledge_conflict_survey, WhoQA} also aim to examine the role of parametric knowledge in LLMs, they are not designed for evaluating LCLMs. To address this gap, we first construct a dataset to investigate whether LCLMs naturally incorporate parametric knowledge during long-context generation across varying context lengths. Specifically, we choose WhoQA~\citep{WhoQA}, a short-form QA dataset that provides multiple context-answer pairs for each entity, as the foundation for our construction.

Specifically, for each entity in WhoQA, we generated an answer to every associated question in the dataset and retained only those entities where the model consistently produced a single, invariant answer. This filtering process ensures that the model’s parametric knowledge encodes a single, unambiguous answer for the retained entities. Since parametric knowledge may differ across LCLMs, we constructed a separate dataset of 300 examples for each model, which we refer to as the \textbf{I-WhoQA} dataset.

To assess how parametric knowledge influences long-context generation, we select two distinct context-answer pairs for each example in the I-WhoQA dataset: (1) a context that aligns with the model’s parametric knowledge, referred to as the \textbf{I-WhoQA-Parametric} subset; (2) a context that conflicts with the model’s parametric knowledge, referred to as the \textbf{I-WhoQA-Conflict} subset.
By comparing model performance across these two subsets and varying context lengths, we aim to reveal the extent to which parametric knowledge impacts long-context generation. An example from the I-WhoQA dataset is provided in the Appendix Figure ~\ref{fig:appendix_iwhoqa}.

\subsection{Experiment}

\paragraph{Experimental Setting} To validate the role of parametric knowledge in long-context generation, we conducted experiments using two curated I-WhoQA subsets mentioned above, alongside two different backend models — \texttt{mistral-7b-instruct-v0.2}~\citep{Mistral} and \texttt{Llama3.1-8B-instruct}~\citep{llama3}. For each backend model, we employed greedy decoding strategy with a maximum of 32 generation tokens. To investigate the effect of context length, a key factor in long-context generation, we report results across varying context lengths. For each example, the only difference across these different context settings is the amount of irrelevant content inserted into the context, which is consistent with the setup of Needle-in-a-Haystack test. In our evaluation, we only include answers that clearly align with one of two competing knowledge sources: the injected context (which can represent either parametric or conflict knowledge) or the alternative source (i.e., the one not injected). This ensures a clean comparison between model behavior in context-following vs. parametric recall scenarios.

\begin{figure*}[t]
    \centering
    \includegraphics[width=0.80\linewidth]{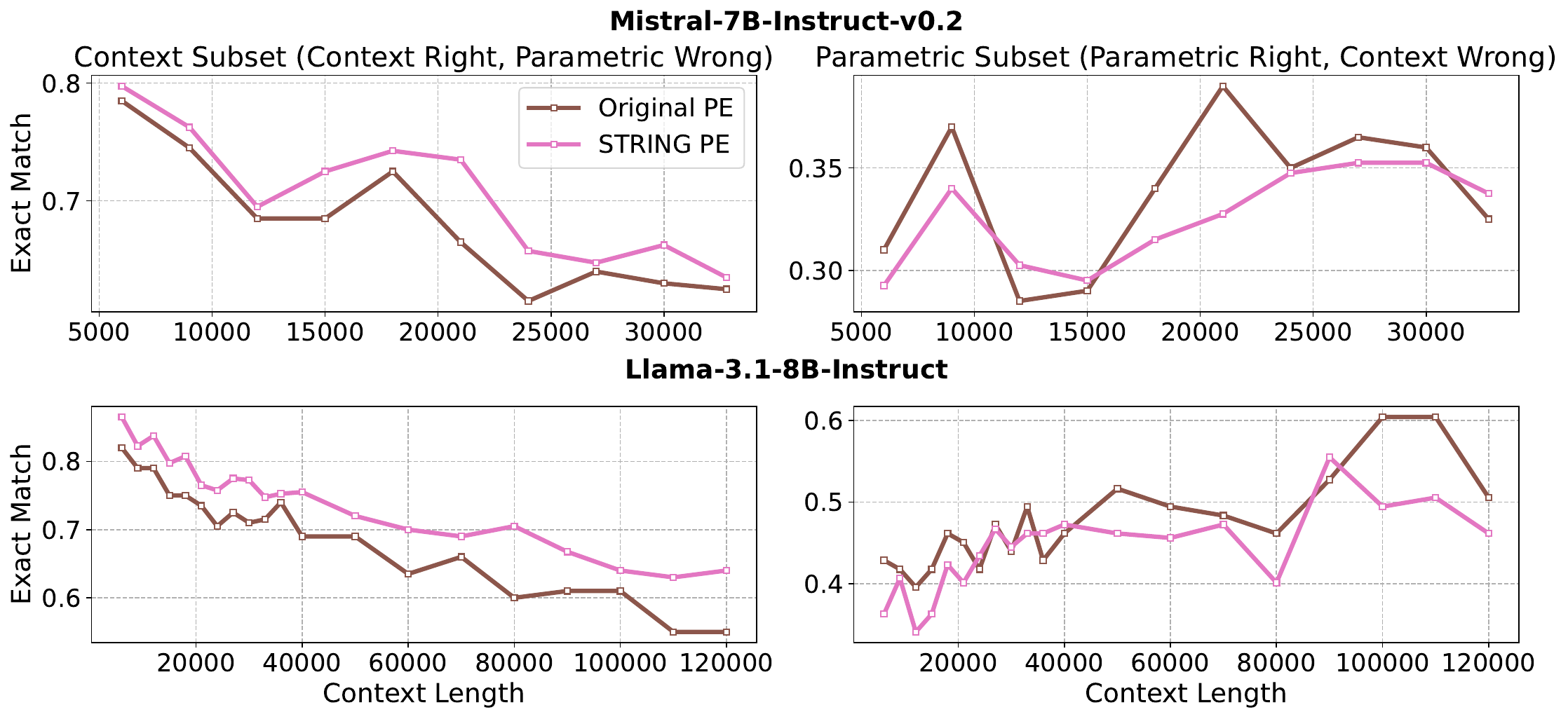}

    \caption{Performance of LCLMs on HotPotQA-Context and -Parametric subsets. Left) STRING improves performance on the HotPotQA-context subset by enhancing extrinsic retrieval ability. Right) STRING hinders the model's ability to recall parametric knowledge and leads to the decrease of performance.
    }
    \label{fig:section2_hotpot}
\end{figure*}

\paragraph{Main Results} Results in Figure~\ref{fig:section1} show that the parametric knowledge of LCLMs influences long-context generation and this influence becomes more pronounced as context length increases. First, the left side of Figure~\ref{fig:section1} (Answer Aligned With the Inserted Knowledge) shows the performance on the IWhoQA-Conflict subset is consistently worse than on IWhoQA-Parametric subset. This demonstrates that LCLMs are struggling to follow the given context when its parametric knowledge contains a different answer. Second, from the right side of Figure~\ref{fig:section1} (Answer Aligned With the Opposing Knowledge) on the I-WhoQA-Conflict subset, where conflicting contexts are provided as supporting facts, we observed a notable increase in the proportion of outputs adhering to opposing knowledge (i. e. parametric knowledge) as context length increased. This trend is especially pronounced when \texttt{Llama-3.1-8B-Instruct} serves as the backend model, which can handle longer context lengths. \textbf{This suggests that as the context length increases, the model increasingly relies on its parametric knowledge even when that knowledge contradicts the external context.}

\section{The Trade-off Between Parametric Recall Ability and Extrinsic Retrieval Ability}

In Section 1, we demonstrated that the parametric knowledge within LCLMs increasingly influences generation as context length grows. While existing benchmarks and research~\citep{string_pe, lm-infinite, DynamicNTK} on LCLMs have largely focused on evaluating and improving extrinsic retrieval, little attention has been paid to how parametric knowledge affects long-context generation. In this section, we aim to answer the question:
\textbf{Can we improve extrinsic retrieval without compromising parametric recall in long-context generation?} We examine whether STRING~\citep{string_pe}, an improved variant of RoPE designed for long-context modeling, can enhance both capabilities. Experimental results reveal that although STRING improves extrinsic retrieval, it can inadvertently impair parametric recall, sometimes leading to a decline in overall model performance.

\begin{figure}[t]
    \centering
    \includegraphics[width=0.89\linewidth]{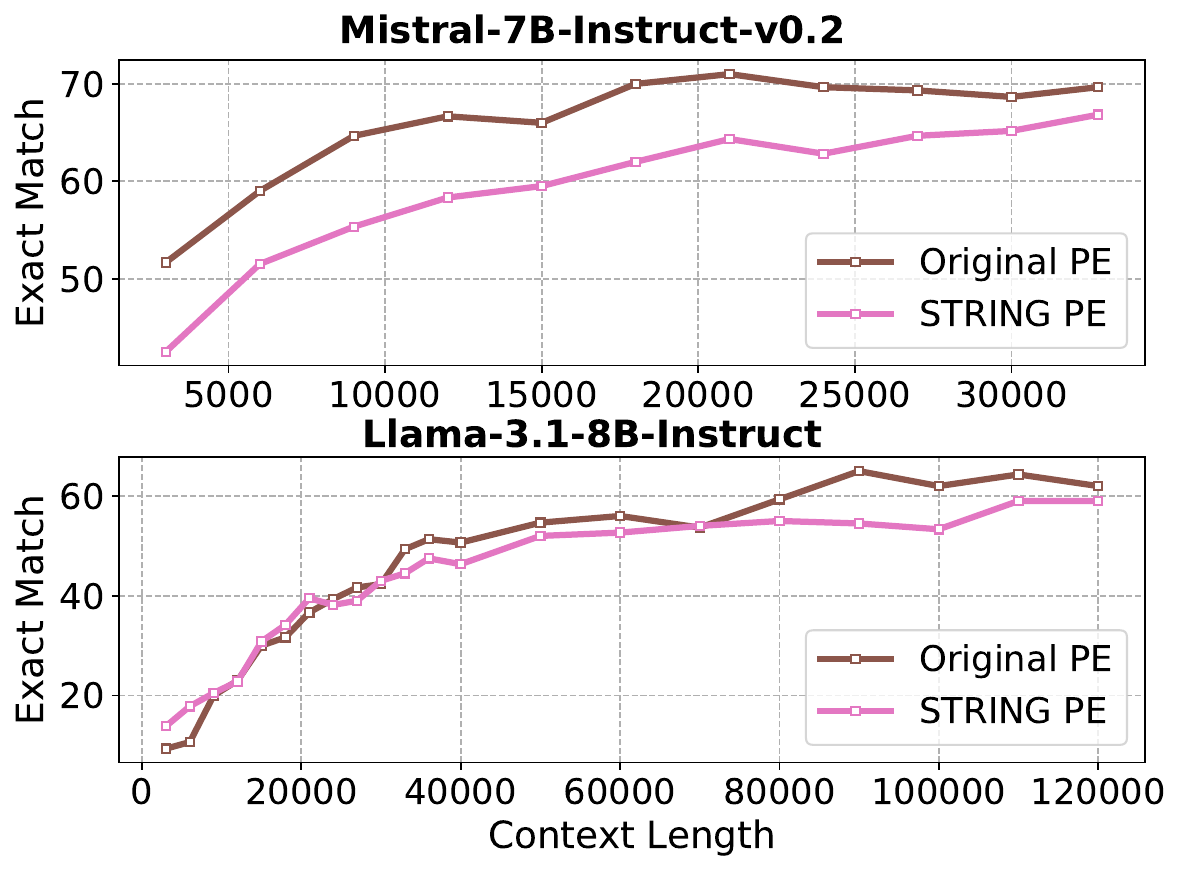}

    \caption{I-WhoQA-Irrelevant subsets: As the length of the context is increased, LCLMs ignore the context and generate according to their parametric knowledge. However, using STRING causes LCLMs to re-focus on irrelevant context and generate wrong answers.}
    \label{fig:section2_irrelevant}
\end{figure}

\begin{figure*}
    \centering
    \includegraphics[width=0.90\linewidth]{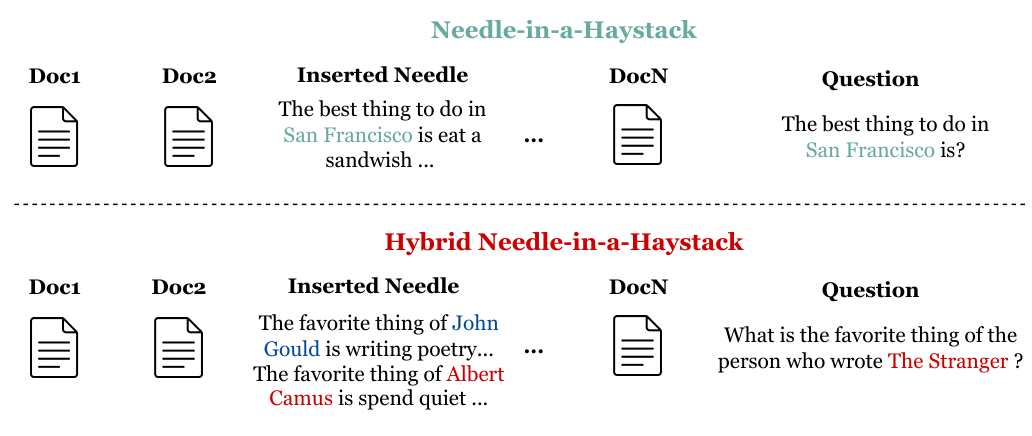}
    \caption{Upper) Needle-in-a-Haystack. It involves directly inserting the answer into the haystack and retrieving it. Lower) Hybrid Needle-in-a-Haystack. It requires a two-step process: first, performing parametric recall to identify the retrieval target based on the model’s parametric knowledge, and then retrieving the answer from the haystack.}
    \label{fig:hybrid_niah}
\end{figure*}
\subsection{Dataset Creation}
In Section 1, we constructed the I-WhoQA dataset, ensuring that all examples can be answered using the model's parametric knowledge. To further assess the trade-off between extrinsic retrieval and parametric recall, we created the \textbf{I-WhoQA-Irrelevant} subset. In this subset, instead of providing context relevant to the current question, we inserted entirely irrelevant content to evaluate how models handle unrelated information, and whether STRING remains effective when answering the question requires parametric knowledge rather than context-based extrinsic retrieval.


In addition to the \textbf{I-WhoQA-Irrelevant} subset, we construct a more realistic dataset based on HotpotQA~\citep{yang2018hotpotqa} to simulate scenarios where the context appears relevant to the question but ultimately fails to support the correct answer. From the HotpotQA dataset, we derive two subsets by analyzing the relationship between the external context, the model’s parametric knowledge, and the reference answer.

First, to distinguish the role of external context from that of the model's parametric knowledge, we ensured that the answers derived from these two sources are not the same. Building on this distinction, we obtained two final subsets. The first subset, referred to as the \textbf{HotpotQA-Context} subset, contains examples where the reference answer can be derived from the given contexts rather than the parametric knowledge. The second subset, known as the \textbf{HotpotQA-Parametric} subset, consists of examples where the reference answer matches the model’s parametric knowledge. We constructed a dataset consisting of 200 examples for each subset, except for the HotpotQA-Parametric subset for \texttt{Llama3.1-8B-Instruct}, which contains a total of 93 examples.

\subsection{Experiment}
\paragraph{Experimental Setting} We use the same backend models and generation configuration as described in Section 1, applied to the newly constructed datasets. For STRING, we configure the local value to 128 and two different shift-ratios: 0.25 and 0.33. We report the average performance across these two settings. More details about STRING can be found in~\citet{string_pe}.

\paragraph{Main Results} 

As shown in Figure~\ref{fig:section2_irrelevant}, STRING consistently underperforms the baseline RoPE method in terms of accuracy. This performance degradation can be attributed to STRING's stronger extrinsic retrieval bias in long-context settings, which causes LCLMs to incorrectly focus on irrelevant context. Even when the external context is entirely unrelated to the question, STRING tends to extract the answer from the context rather than leveraging the model’s parametric knowledge, which actually contains the correct answer. 

Results on HotpotQA-Context subset and HotpotQA-Parametric subset are shown in Figure~\ref{fig:section2_hotpot}. STRING significantly improves performance on the HotpotQA-Context subset, where the model must rely on the external context to derive the correct answer.  This confirmed that STRING is improving extrinsic retrieval ability on LCLMs. However, on the HotpotQA-Parametric subset where the context is related but not useful, STRING underperforms RoPE and this is consistent with our finding on the I-WhoQA-Irrelevant subset. \textbf{In conclusion, STRING's enhanced extrinsic retrieval ability interferes with the model’s use of parametric knowledge, leading to degraded performance in scenarios where the context is misleading or insufficient.} This trade-off becomes more pronounced as the context length increases since STRING's boosted extrinsic retrieval ability can be more prominent with longer contexts as shown on the \texttt{Llama-3.1-8B-Instruct} HotPotQA-Context subset.

\begin{table*}
\begin{center}
\resizebox{0.80\linewidth}{!}{
\begin{tabular}{ccccccccc}
\hline
\multicolumn{1}{c|}{\multirow{3}{*}{Model}}      & \multicolumn{4}{c|}{Generation Length = 32}         & \multicolumn{4}{c}{Generation Length = 64} \\ \cline{2-9} 
\multicolumn{1}{c|}{}                            & \multicolumn{4}{c|}{Random Facts}                  & \multicolumn{4}{c}{Random Facts}           \\ \cline{2-9} 
\multicolumn{1}{c|}{}                            & 0     & 1     & 2     & \multicolumn{1}{c|}{3}     & 0         & 1        & 2        & 3        \\ \hline
\multicolumn{9}{c}{\textbf{Needle-in-a-Haystack}}                                                                                                         \\ \hline
\multicolumn{1}{c|}{Mistral-7B-Instruct-v0.2}    & 100   & 100   & 99.55 & \multicolumn{1}{c|}{98.41} & -         & -        & -        & -        \\
\multicolumn{1}{c|}{Llama-3.1-8B-Instruct}       & 100   & 99.21 & 99.33 & \multicolumn{1}{c|}{98.83} & -         & -        & -        & -        \\
\multicolumn{1}{c|}{Qwen2.5-7B-Instruct}         & 99.78 & 99.64 & 99.81 & \multicolumn{1}{c|}{99.92} & -         & -        & -        & -        \\ \hline
\multicolumn{9}{c}{\textbf{Hybrid Needle-in-a-Haystack}}                                                                                                  \\ \hline
\multicolumn{1}{c|}{Mistral-7B-Instruct-v0.2}    & 89.31 & 58.30 & 53.53 & \multicolumn{1}{c|}{53.21} & 97.83     & 63.28    & 58.16    & 58.11    \\
\multicolumn{1}{c|}{Mistral-7B-Instruct-v0.3}    & 73.17 & 59.54 & 56.97 & \multicolumn{1}{c|}{56.94} & 76.56     & 62.48    & 58.01    & 57.73    \\ \hline
\multicolumn{1}{c|}{Llama-3.1-8B-Instruct}       & 92.97 & 83.66 & 72.60 & \multicolumn{1}{c|}{67.47} & 96.71     & 90.28    & 80.93    & 74.39    \\
\multicolumn{1}{c|}{Llama-3.1-70B-Instruct}      & 95.73 & 86.39 & 74.48 & \multicolumn{1}{c|}{71.81} & 95.77     & 89.88    & 77.66    & 74.68    \\ \hline
\multicolumn{1}{c|}{Qwen2.5-7B-Instruct}         & 86.38 & 83.32 & 77.23 & \multicolumn{1}{c|}{73.14} & 90.72     & 88.46    & 80.98    & 77.03    \\
\multicolumn{1}{c|}{Qwen2.5-14B-Instruct}        & 54.46 & 76.74 & 75.32 & \multicolumn{1}{c|}{73.97} & 92.95     & 94.32    & 89.85    & 85.60    \\
\multicolumn{1}{c|}{Qwen2.5-72B-Instruct}        & 64.01 & 98.13 & 97.92 & \multicolumn{1}{c|}{97.77} & 94.16     & 99.53    & 99.84    & 98.86    \\
\multicolumn{1}{c|}{Qwen2.5-7B-Instruct-1M-128k} & 97.72 & 84.83 & 82.10 & \multicolumn{1}{c|}{79.31} & 98.50     & 84.89    & 82.22    & 79.37    \\
\multicolumn{1}{c|}{Qwen2.5-7B-Instruct-1M}      & 60.07 & 48.09 & 43.70 & \multicolumn{1}{c|}{42.91} & 60.77     & 49.26    & 43.43    & 43.21        \\ \hline
\end{tabular}
}
\end{center}
\caption{Average accuracy score of different models. \texttt{Qwen2.5-7B-Instruct-1M-128k} means we set the max input length to 128k.}
\label{table:main}
\end{table*}

\section{Hybrid Needle-in-a-Haystack Framework}

So far, we have shown that parametric knowledge plays a crucial role in the long-context generation, yet improvements in extrinsic retrieval do not translate into better parametric recall. Moreover, existing benchmarks~\citep{hsieh2024_RULER, an2023leval, infinity-bench} primarily focus on how to utilize external context, overlooking scenarios that require interaction with parametric knowledge. As a result, methods~\citep{string_pe} developed under these benchmarks fail to achieve better performance when parametric recall is essential. \textbf{To address this gap, we propose a simple yet effective Hybrid Needle-in-a-Haystack (NIAH) test designed to jointly evaluate parametric recall ability and extrinsic retrieval ability.} This test enables a more comprehensive assessment of how these two forms of knowledge interact, complement, or interfere with one another, ultimately affecting the quality of model generation.

\subsection{Dataset Creation}

NIAH test uses irrelevant content as a haystack and inserts the answer from a question-answer pair into the haystack to evaluate LCLMs' extrinsic retrieval ability under different context lengths and insertion depths. Unlike the standard NIAH test, our Hybrid NIAH test simultaneously assesses parametric recall ability and extrinsic retrieval ability by designing scenarios where the model must first utilize its parametric knowledge and then perform extrinsic retrieval on the haystack. A comparison between NIAH test and Hybrid NIAH test is illustrated in Figure~\ref{fig:hybrid_niah}. The Hybrid NIAH test introduces an additional step designed to trigger the parametric recall ability, requiring the model to recall the relevant target from parametric knowledge. The model relies on this recalled target entity to guide extrinsic retrieval from the external context, ultimately producing the final answer.

Firstly, to implement the parametric recall step, we use the I-WhoQA dataset constructed in Section 1. To ensure that the question-answer pairs only require knowledge that exists across all models, we perform an intersection operation between the I-WhoQA datasets obtained from \texttt{Mistral-7B-Instruct-v0.2} and \texttt{Llama3.1-8B-Instruct}. Secondly, for the extrinsic retrieval step, we adopt a similar approach to the NIAH test and create various random facts to serve as inserted answers to the corresponding questions. For the remaining irrelevant haystack construction, we follow the NIAH test and use the PaulGrahamEssays dataset to construct the target length haystack. 


To prevent the model from exploiting potential syntactic patterns present in the question-answer pairs to extract the inserted ``needle'' from the context, we introduce varying amounts of random facts as interference (from 0 to 3). This additional noise ensures that models understand the task and the context rather than relying on superficial patterns, thereby enhancing the robustness and reliability of the final result. Details of the final knowledge pairs are listed in the Appendix Section 1.

\subsection{Experiment}

\paragraph{Experimental Setting} 
We evaluated models from three different families with various specifications: Mistral, Llama3.1, and Qwen2.5. For all models, we utilized the Instruct versions and employed greedy decoding strategy. We set the maximum generation length to 32 and 64.  For each model, we divided the maximum context length into 40 intervals. For each interval, we defined 10 progressively increasing insertion depths and conducted Hybrid NIAH tests accordingly. This setup allows us to systematically evaluate the models’ performance across varying context lengths and insertion depths. To ensure the stability of the results, we use multiple examples and report the average result. Finally, to measure the similarity between the predictions and the references, we adopt the method provided by PyramidKV \citep{cai2024pyramidkv}, which calculates the final result based on exact matches between tokens. The formula is as follows:
$$
Score = \frac{|P| \cap |A|}{|A|}
$$
where $P$ and $A$ are tokens from predictions and references respectively. More detail settings are listed in the Appendix Table~\ref{table:appendix_setup}.


\begin{figure*}[t]
    \centering
    \includegraphics[width=0.94\linewidth]{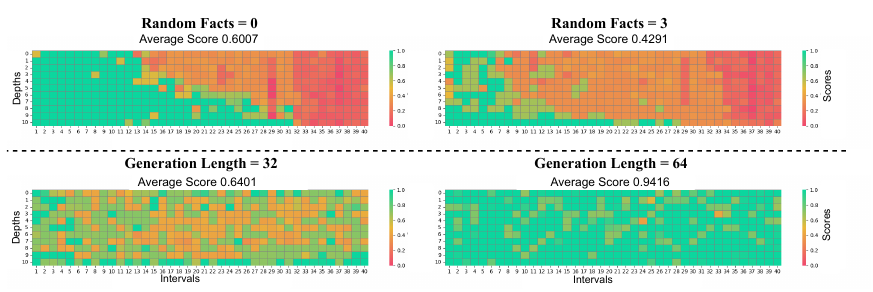}
    \caption{Hybrid NIAH test results. Upper) \texttt{Qwen2.5-7B-Instruct-1M} with generation length 32 and various numbers of inserted random facts. Lower) \texttt{Qwen2.5-72B-Instruct} with generation lengths 32 and 64. The number of random facts was set to 0.}
    \label{fig:niah_figures}
\end{figure*}

\paragraph{Main Results}

The averaged results for the Hybrid NIAH test are presented in Table~\ref{table:main} and we list the detailed results in the Appendix Section 2. From these results, we can observe three interesting findings. 

Firstly, with only a simple modification to the question, our Hybrid NIAH test becomes significantly more challenging than the standard NIAH test, as shown in Table~\ref{table:main}. This effect is especially evident in long-context scenarios, as demonstrated by the results on \texttt{Qwen2.5-7B-Instruct-1M} in Figure~\ref{fig:niah_figures}. Moreover, by inserting varying amounts of random facts, we observe a clear decline in model performance by up to 25\% for \texttt{Llama-3.1-70B-Instruct}, while the NIAH test results remain unaffected by this interference, as shown in Table~\ref{table:main}. This discrepancy indicates that models rely on superficial patterns within the context to extract answers when there is a single needle, rather than genuinely leveraging their parametric knowledge. The insertion of random facts effectively disrupts this pattern-based retrieval process, allowing us to better assess whether a model truly utilizes its parametric knowledge rather than merely exploiting contextual shortcuts. Multiple-Needles Hybrid NIAH test serves as a more rigorous evaluation of how well models integrate their parametric recall ability with the long-context generation.


Secondly, models from the Mistral and Llama3.1 families fail to achieve significant improvements on the Hybrid-NIAH test, even when upgraded to larger versions. This suggests that these models struggle to effectively utilize their parametric knowledge under long-context scenarios. Despite increased model capacity, their performance remains constrained when external context is incomplete or ambiguous. In contrast, the Qwen2.5 family shows substantial improvements as model size scales up. Larger Qwen2.5 models are more capable of effectively leveraging their parametric knowledge, providing more accurate and factually consistent results even when the external context is incomplete.

Finally, we observe an interesting phenomenon in the Qwen2.5 models for Single-Needle Hybrid NIAH test: larger model variants initially claim to not see the inserted needle but are able to generate and then retrieve it eventually. As shown in the Appendix Figure~\ref{fig:appendix_niah_qwen2.5}, \texttt{Qwen2.5-14B-Instruct} and \texttt{Qwen2.5-72B-Instruct} start by generating a refusal but then successfully retrieve the needle regardless. In contrast, the smaller \texttt{Qwen2.5-7B-Instruct} directly retrieves the needle without any refusal. This behavior explains the findings in Table~\ref{table:main} where \texttt{Qwen2.5-14B-Instruct} and \texttt{Qwen2.5-72B-Instruct} underperforming \texttt{Qwen2.5-7B-Instruct} when generation length is 32. Limiting the generation length only allows the larger Qwen models to generate the initial refusal and not the retrieved needle. Strangely, this refusal behavior is not observed when there are multiple inserted needles. Since this behavior is unique to the Qwen2.5 family and only for single Hybrid NIAH test, we hypothesize it is related to the Qwen's chunked attention mechanism \citep{dual_chunk_attention} struggling to effectively focus on the singular inserted needle when generating the first token. When multiple needles are inserted, the syntactical similarity between the needles helps the larger Qwen models to focus on the context and they do not refuse to answer initially. We leave further investigation of this phenomenon for future work.

\section{Conclusion}

In this work, we investigate the role of parametric knowledge in LCLMs and uncover a critical trade-off between parametric recall and extrinsic retrieval. Our analysis reveals that enhancements achieved on extrinsic retrieval ability can inadvertently suppress the model's use of its parametric knowledge, especially when the context is irrelevant or misleading. To evaluate this interplay, we introduce the novel Hybrid Needle-in-a-Haystack test that assesses a model’s ability to jointly integrate parametric and extrinsic knowledge. Experimental results demonstrate that even large-scale LCLMs struggle to effectively combine these two abilities and often fail to fully leverage their parametric knowledge during long-context generation. Our findings highlight the need for future model designs and evaluation protocols that consider both parametric and extrinsic knowledge sources, paving the way for more robust and context-aware LCLMs.

\bibliography{aaai2026}


\clearpage
\appendix

\begin{figure*}[t]
    \centering
    \includegraphics[width=0.75\linewidth]{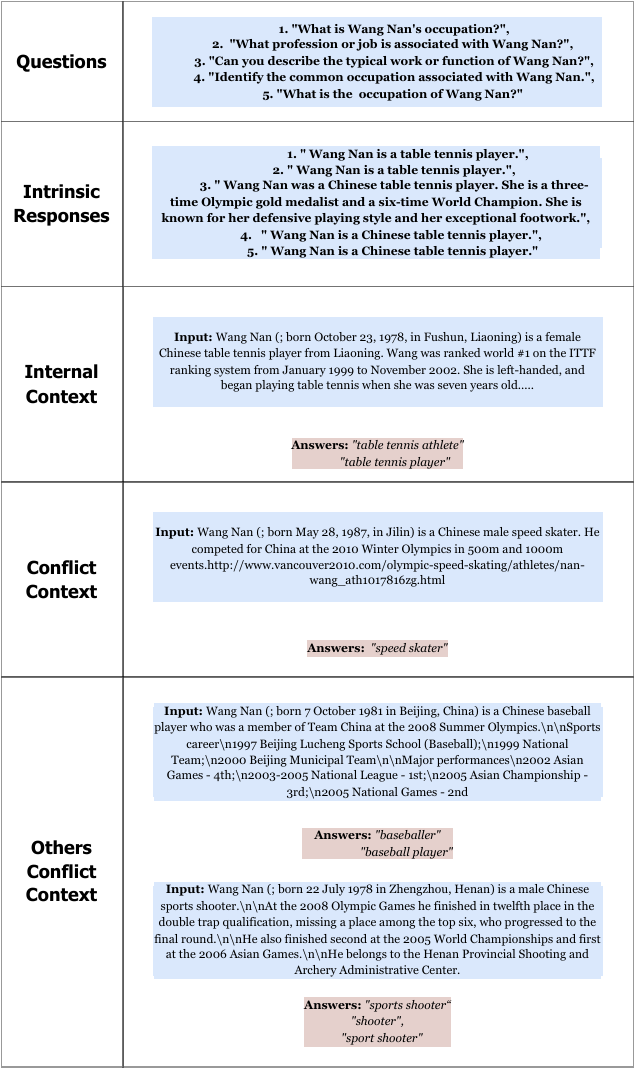}
    \caption{One example on I-WhoQA dataset on \texttt{Mistral-7B-Instruct-v0.2}.}
    \label{fig:appendix_iwhoqa}
\end{figure*}
\clearpage

\begin{table*}[]
\resizebox{\linewidth}{!}{
\begin{tabular}{ccccc}
\hline
Model                       & Num of Examples & Max Context Length & Num of Depths & Num of Intervals \\ \hline
Mistral-7B-Instruct-v0.2    & 5               & 32768              & 40            & 10               \\
Mistral-7B-Instruct-v0.3    & 2               & 32768              & 40            & 10               \\
Llama-3.1-8B-Instruct       & 5               & 131072             & 40            & 10               \\
Llama-3.1-70B-Instruct      & 2               & 131072             & 40            & 10               \\
Qwen2.5-7B-Instruct         & 5               & 131072             & 40            & 10               \\
Qwen2.5-14B-Instruct        & 2               & 131072             & 40            & 10               \\
Qwen2.5-72B-Instruct        & 2               & 131072             & 40            & 10               \\
Qwen2.5-7B-Instruct-1M-128k & 5               & 131072             & 40            & 10               \\
Qwen2.5-7B-Instruct-1M      & 2               & 1010000            & 40            & 10               \\ \hline
\end{tabular}
}
\caption{NIAH and Hybrid NIAH test configurations for different backend models.}
\label{table:appendix_setup}
\end{table*}

\begin{figure*}
    \centering
    \includegraphics[width=\linewidth]{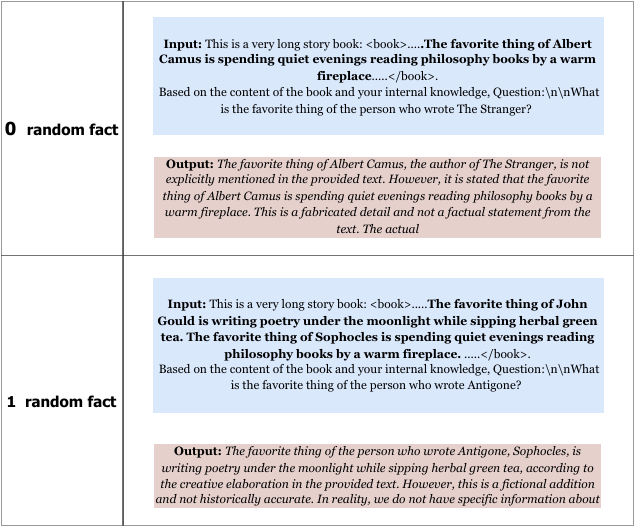}
    \caption{Hybrid NIAH examples from Qwen2.5-72B-Instruct. The generation length was set to 64.}
    \label{fig:appendix_niah_example}
\end{figure*}

\section{Hybrid Needle-in-a-Haystack Knowledge Dataset}
\label{sections:appendix_knowledges}

In the WhoQA dataset, various entity types are provided, including [date of birth], [author], [country], [date of death], [father], [composer], [performer], [genre], [creator], [occupation], [spouse], [publisher], and [mother]. For constructing the question-answer pairs used in Hybrid NIAH test, we focus specifically on the [author] entity type as the basis for our evaluations.

All entities used in Hybrid NIAH tests are illustrated in Table ~\ref{fig:appendix_knowledges}, with the bolded entries representing the target entities used for generating the desired answers. To construct the random facts required for our Hybrid NIAH tests, we randomly select entities from all entities other than the current target entity.

\begin{table*}[]
\resizebox{\linewidth}{!}{
\begin{tabular}{cc|cc}
\hline
Book                            & Author                 & Book                       & Author                \\ \hline
\textbf{The Stranger}           & \textbf{Albert Camus}  & The Social Contract        & Jean-Jacques Rousseau \\
\textbf{Antigone}               & \textbf{Sophocles}     & Crime and Punishment       & Fyodor Dostoevsky     \\
\textbf{Leviathan}              & \textbf{Thomas Hobbes} & Iphigenia in Tauris        & Euripides             \\
\textbf{Dracula}                & \textbf{Bram Stoker}   & Dover Beach                & Matthew Arnold        \\
\textbf{The Birds of Australia} & \textbf{John Gould}    & Travels with My Aunt       & Graham Greene         \\
Conan the Barbarian             & Robert E. Howard       & Candide                    & Voltaire              \\
Politics                        & Aristotle              & Time's Arrow               & Martin Amis           \\
What's the Matter with Kansas?  & Thomas Frank           & Bad Science                & Ben Goldacre          \\
On Liberty                      & John Stuart Mill       & A Clash of Kings           & George R. R. Martin   \\
Tipping the Velvet              & Sarah Waters           & Shakuntala                 & Kalidasa              \\
The Tale of Genji               & Murasaki Shikibu       & The Age of Reason          & Thomas Paine          \\
The Art of War                  & Sun Tzu                & Ulysses                    & James Joyce           \\
The Prince                      & Niccolò Machiavelli    & Lady Lazarus               & Sylvia Plath          \\
Carmen                          & Georges Bizet          & The Age of Innocence       & Edith Wharton         \\
The Temple of Elemental Evil    & Gary Gygax             & James Bond                 & Ian Fleming           \\
Histories                       & Herodotus              & The Fishermen              & Chigozie Obioma       \\
Pygmalion                       & George Bernard Shaw    & The Three Musketeers       & Alexandre Dumas       \\
The Hiding Place                & Corrie ten Boom        & The Wonderful Wizard of Oz & L. Frank Baum         \\
Misery                          & Stephen King           & Richard III                & William Shakespeare   \\
The Nine Billion Names of God   & Arthur C. Clarke       & Amazing Grace              & John Newton           \\
The Seagull                     & Anton Chekhov          & Paradise Lost              & John Milton           \\
There Will Come Soft Rains      & Ray Bradbury           & The Betrothed              & Alessandro Manzoni    \\
Parable of the Sower            & Octavia Butler         & The Hollow Men             & T.S. Eliot            \\ \hline
\end{tabular}
}
\caption{Knowledge pairs selected from I-WhoQA dataset.}
\label{fig:appendix_knowledges}
\end{table*}

\section{Hybrid Needle-in-a-Haystack Results}
\label{sections:appendix_niah_details}

\begin{figure*}
    \centering
    \includegraphics[width=\linewidth]{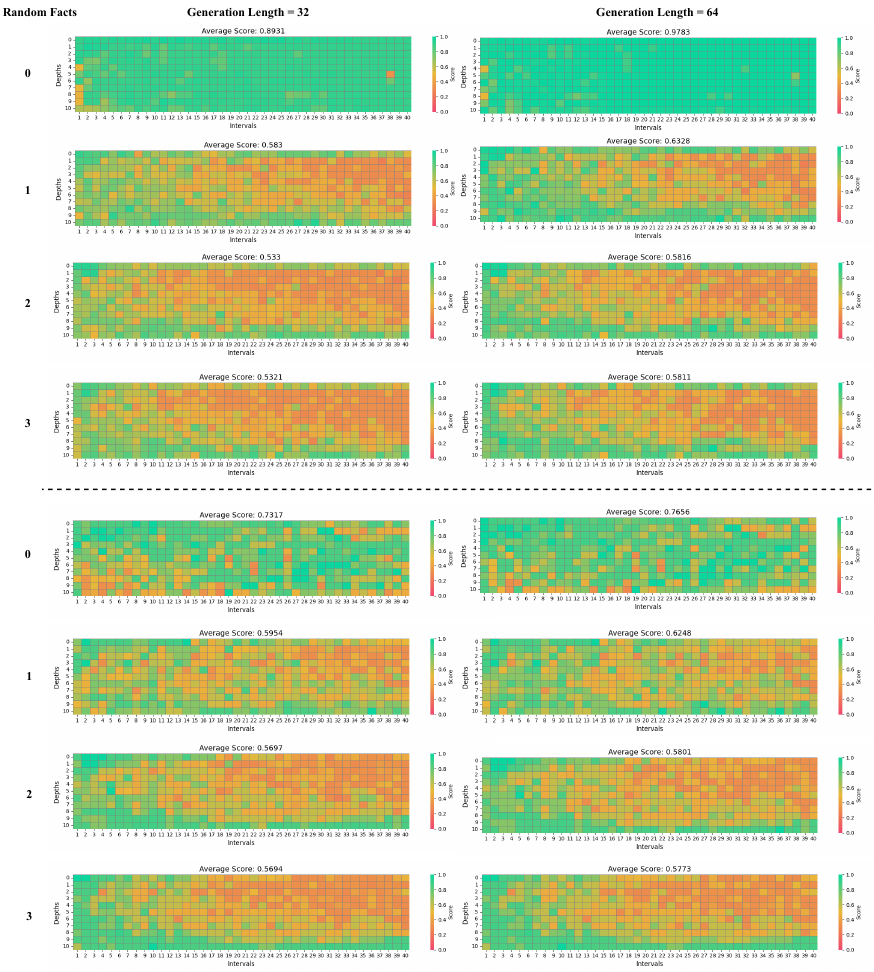}
    \caption{Hybrid Needle-in-a-Haystack test results on models from Mistral family. Upper) \texttt{Mistral-7B-Instruct-v0.2} and Lower) \texttt{Mistral-7B-Instruct-v0.3}.}
    \label{fig:appendix_niah_mistral}
\end{figure*}

\begin{figure*}
    \centering
    \includegraphics[width=1.05\linewidth]{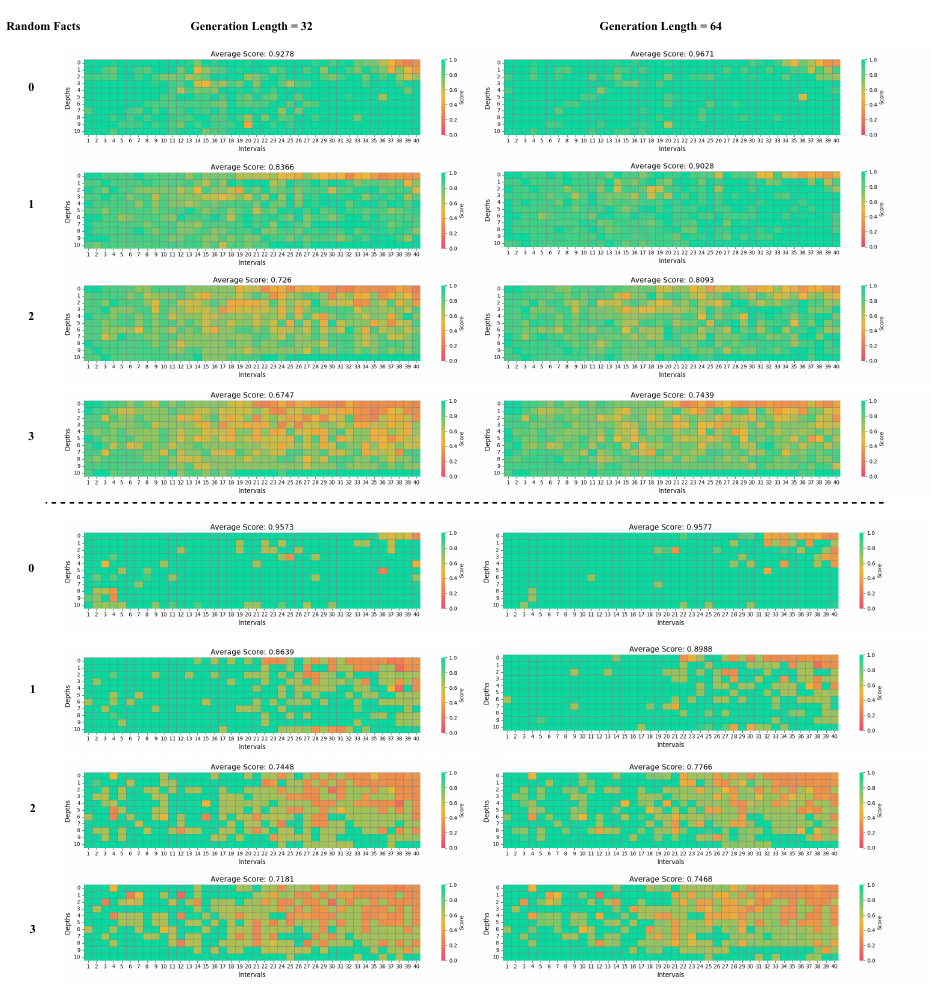}
    \caption{Hybrid Needle-in-a-Haystack test results on models from Llama3.1 family. Upper) \texttt{Llama3.1-8B-Instruct} and Lower) \texttt{Llama3.1-70B-Instruct}.}
    \label{fig:appendix_niah_llama3}
\end{figure*}

\begin{figure*}
    \centering
    \includegraphics[width=1.05\linewidth]{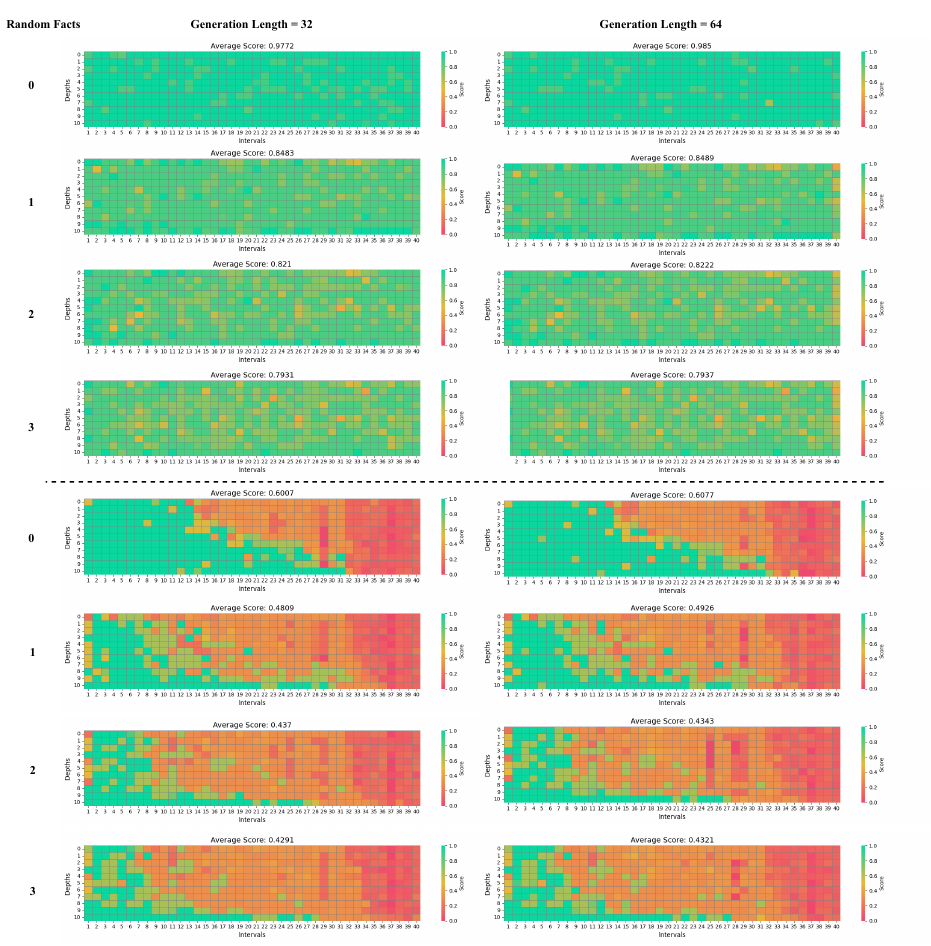}
    \caption{Hybrid Needle-in-a-Haystack test results on \texttt{Qwen2.5-7B-Instruct-1M}. Upper) \texttt{Qwen2.5-7B-Instruct-1M-128k}, Lower) \texttt{Qwen2.5-7B-Instruct-1M}.}
    \label{fig:appendix_niah_qwen2.5_1m}
\end{figure*}

We listed Hybrid NIAH test results on different models. Figure~\ref{fig:appendix_niah_mistral} shows the test results on models from Mistral family. Figure~\ref{fig:appendix_niah_llama3} shows the test results on models from Llama3.1 family. Figure~\ref{fig:appendix_niah_qwen2.5_1m} shows the test results on models from Qwen2.5 family except \texttt{Qwen2.5-7B-Instruct-1M}. Figure~\ref{fig:appendix_niah_qwen2.5_1m} shows the test results on \texttt{Qwen2.5-7B-Instruct-1M}.

\begin{figure*}
    \centering
    \includegraphics[width=0.9\linewidth]{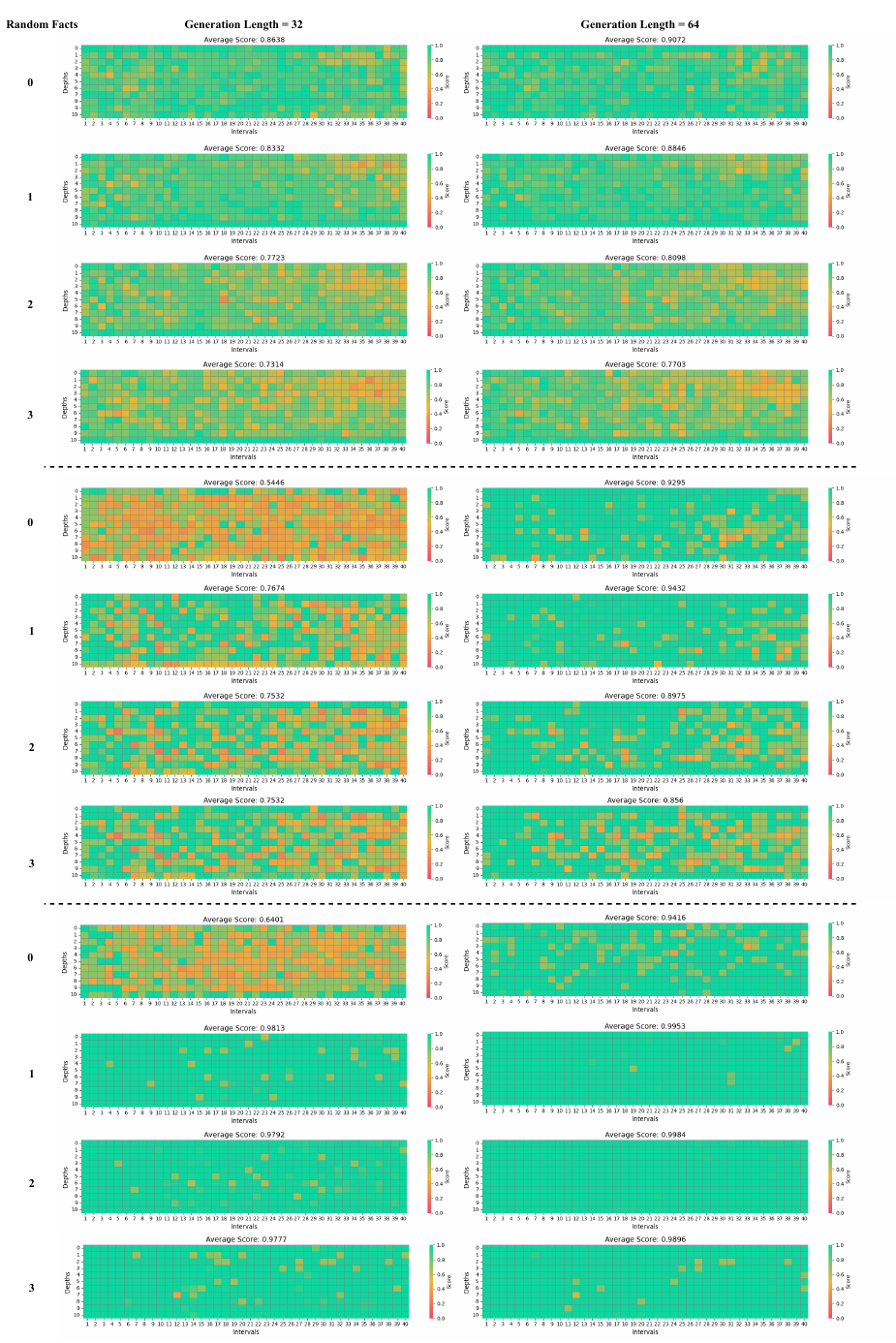}
    \caption{Hybrid Needle-in-a-Haystack test results on models from Qwen2.5 family. Upper) \texttt{Qwen2.5-7B-Instruct}, Middle) \texttt{Qwen2.5-14B-Instruct} Lower) \texttt{Qwen2.5-70B-Instruct}.}
    \label{fig:appendix_niah_qwen2.5}
\end{figure*}

\end{document}